\documentclass[sigconf]{acmart}
\usepackage{tabularx}

\AtBeginDocument{%
  }

\setcopyright{none}
\copyrightyear{2026}
\acmYear{2026}
\acmDOI{}
\acmConference[CIKM '26]{ACM International Conference on Information and Knowledge Management}{2026}{TBD}
\acmISBN{}
\renewcommand\footnotetextcopyrightpermission[1]{}

\begin{document}

\title{Object-Centric Dataset Resources for Constrained-Data Image Generation and Augmentation}

%%
%% The authors list 

\author{Vasile Marian}
\affiliation{%
  \institution{The University of Queensland}
  \city{Brisbane}
  \country{Australia}}
\email{vasile.marian@student.uq.edu.au}

\author{Yong-Bin Kang}
\affiliation{%
  \institution{Swinburne University of Technology}
  \city{Melbourne}
  \country{Australia}}
\email{ykang@swin.edu.au}

\author{Alexander Buddery}
\affiliation{%
  \institution{The University of Queensland}
  \city{Brisbane}
  \country{Australia}}
\email{alexander.buddery@cookmedical.com}

%% headers. Often, this list is too long and will overlap
%% other information printed in the page headers.
\renewcommand{\shortauthors}{Marian et al.}

\begin{abstract}
Object-centric image generation is important for applications where additional labelled examples are difficult to collect, such as pedestrian analysis in smart-city scenes, traffic-sign inspection, and domain-specific object detection. In these settings, synthetic images are most useful for training and evaluation when datasets preserve object structure, bounding-box information, visual diversity, and realistic context. Existing image datasets are usually designed primarily for classification, detection, or scene understanding rather than for controlled object-centric image generation and augmentation under limited class-specific data. We present a shareable collection of three object-centric dataset resources: Cityscapes--Pedestrian, TrafficSigns, and COCO PottedPlant. The collection standardizes $256\times256$ object-centric crops and bounding-box annotations across three complementary regimes: dense pedestrian scenes with privacy blur and occlusion, cleaner high-contrast traffic signs, and context-diverse potted-plant scenes. The current release contains 3,009 TrafficSigns samples, 2,156 Cityscapes--Pedestrian manifest records, and 7,679 COCO PottedPlant manifest records. The larger COCO-derived manifest is retained to preserve contextual and multi-instance diversity. Users can still draw equal-size subsets with a fixed random seed when controlled comparisons are needed. The release provides direct TrafficSigns data where redistribution is permitted, and scripts, manifests, box-level annotation tables, checksums, and reconstruction documentation for the Cityscapes- and COCO-derived subsets. The release is available through \href{https://github.com/Latzi/object-centric-low-data-datasets}{GitHub} and archived on Zenodo with DOI \href{https://doi.org/10.5281/zenodo.20573001}{10.5281/zenodo.20573001}. It is designed to help users inspect labels and splits, create equal-size subsets, rerun reconstruction scripts where upstream data are available, and evaluate object-centric image generation or synthetic-data augmentation methods on the same records.
\end{abstract}

%%
%% ACM CCS concepts.
%% These terms are selected for the dataset/resource, computer vision,
%% and machine learning focus of the paper.
%%
%%\ccsdesc[500]{Computing methodologies~Computer vision}
%%\ccsdesc[300]{Computing methodologies~Machine learning}
%%\ccsdesc[300]{Information systems~Data analytics}
\ccsdesc[500]{Computing methodologies~Computer vision}
\ccsdesc[300]{Computing methodologies~Image representations}
\ccsdesc[300]{Computing methodologies~Machine learning}
%%
%% Keywords. Separate the keywords with commas.
\keywords{object-centric datasets, constrained-data generation, low-data generation, image generation, dataset resource, synthetic data augmentation} 

\maketitle

\section{Introduction}

Object-centric image generation aims to synthesize images in which a target object remains visually recognizable, spatially grounded, and useful for downstream learning. This is important in data augmentation settings where collecting labelled examples is difficult, including pedestrian examples under occlusion and privacy blur, traffic signs with reliable bounding boxes, and potted plants across varied backgrounds. In these cases, synthetic images are useful only if they preserve object structure, location, and realistic context.

In this paper, \emph{object-centric image generation} means synthesizing images centred on target objects. \emph{Object-centric generative modelling} refers to the models and evaluation protocols used for that task. We use \emph{low-data}, \emph{limited-data}, and \emph{constrained-data} for settings where usable examples, annotation rights, or redistribution permissions are limited compared with broad image corpora and large generative training sets. These terms do not require identical record counts across all subsets. For example, the larger COCO PottedPlant manifest is retained to preserve contextual and multi-instance diversity, while users can draw equal-size subsets with a fixed random seed when needed.

Large-scale vision resources have shaped different parts of this landscape: ImageNet established large-scale supervised image classification \cite{imagenet}, COCO provided object-level annotations for detection and segmentation in complex scenes \cite{coco}, and Cityscapes supported urban scene understanding with dense street-scene annotations \cite{cityscapes}. However, these resources were not primarily designed as constrained-data object-centric generative benchmarks. On the modelling side, GANs introduced adversarial image generation \cite{gan}, while denoising diffusion models established a different generative paradigm based on iterative denoising \cite{ddpm}. Work on data-efficient generation has addressed limited-data training through adaptive discriminator augmentation \cite{stylegan2ada}, differentiable augmentation \cite{diffaug}, and few-shot GAN training strategies \cite{fastgan}.

Constrained object-centric generation exposes several failure modes. A pedestrian generator trained on limited privacy-blurred data may lose body structure or reproduce blur artifacts. A traffic-sign generator may overfit to a few sign appearances rather than learning shape and boundary variation. A potted-plant generator may preserve the object but collapse background diversity. These failures reduce the value of synthetic images for downstream augmentation and evaluation.

A practical problem is that existing datasets are rarely curated for \emph{constrained-data object-centric image generation}. Conditional image-to-image methods such as Pix2Pix show how paired supervision can guide generation \cite{pix2pix}. Layout2Im demonstrates image generation from explicit object layouts \cite{layout2im}, while LayoutDiffusion extends layout-conditioned generation to diffusion-based synthesis \cite{layoutdiffusion}. These methods illustrate the value of spatial supervision, but ad hoc subsets from larger corpora still vary in class mapping, crop construction, centering, filtering, and split definition. Such choices can substantially alter task difficulty, making it harder to reproduce prior work or compare methods fairly.

We address this gap by releasing a shareable collection of three class-specific object-centric dataset resources: Cityscapes--Pedestrian, TrafficSigns, and COCO PottedPlant. The collection provides standardized $256\times256$ crops, bounding-box annotations, sample-level manifests, box-level annotation tables, reconstruction scripts, checksums, and documentation. TrafficSigns is released directly where redistribution is permitted, while the Cityscapes- and COCO-derived subsets are released through reconstruction assets and public metadata.

\paragraph{Relation to dataset resources.}
The release follows established dataset-resource practices by separating source data, annotations, documentation, and provenance. COCO-Stuff extends COCO with additional stuff and thing annotations while keeping the underlying image corpus conceptually distinct from the added annotation layer \cite{cocostuff}. COCO-Text similarly adds text detection and recognition annotations on top of COCO imagery \cite{cocotext}. In the same spirit, our Cityscapes--Pedestrian and COCO PottedPlant subsets release public manifests, per-box annotation tables, scripts, and metadata while respecting upstream image redistribution constraints. The broader release design follows FAIR data principles, which emphasize findability, accessibility, interoperability, and reuse \cite{fair}. It is also informed by Datasheets-for-Datasets, which argues for explicit documentation of dataset motivation, composition, collection, processing, uses, and maintenance \cite{datasheets}.

\paragraph{Contributions.}
This paper makes five contributions. First, it releases three object-centric dataset resources covering dense pedestrian scenes, cleaner traffic-sign crops, and context-diverse potted-plant scenes. Second, it defines a common object-centric benchmark interface based on $256\times256$ crops, bounding boxes, manifests, and split metadata. Third, it provides reproducible construction pipelines covering detection or annotation lookup, box extraction or conversion, class harmonization, cropping, filtering, and metadata export. Fourth, it reports release-derived statistics and qualitative analysis describing object density, multi-box structure, occlusion, blur, object isolation, and contextual diversity. Fifth, it provides a public release package with direct TrafficSigns data where permitted, reconstruction scripts for restricted subsets, manifests with one row per sample, annotation tables with one row per box, fixed-seed subset sampling, checksum verification, processing documentation, and responsible-use notes.

\section{Dataset Construction Pipeline}

\begin{figure*}[t]
  \centering
  \includegraphics[width=0.86\textwidth]{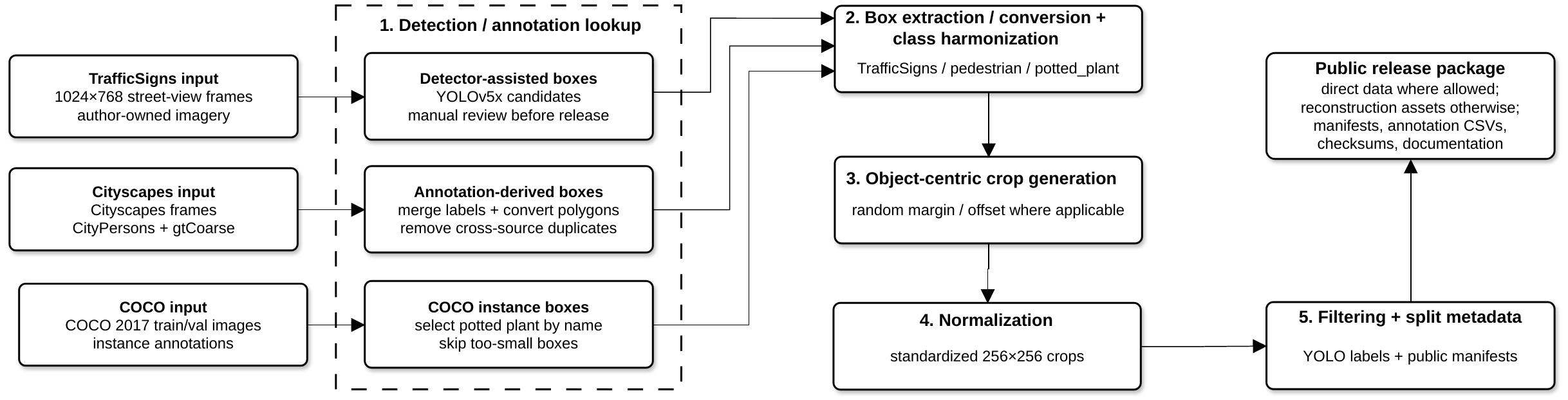}
\caption{Five-stage object-centric curation pipeline. Heterogeneous box sources are standardized into $256\times256$ crops, split metadata, sample-level manifests, box-level annotation tables, checksums, direct data where allowed, and reconstruction assets otherwise.} 
  \Description{Pipeline diagram showing source data from TrafficSigns, Cityscapes, and COCO. The five numbered stages show detection or annotation lookup, bounding-box extraction or conversion with class harmonization, object-centric crop generation, normalization to 256 by 256, filtering and split metadata, and public release artifacts including direct data, manifests, per-box annotation CSVs, scripts, checksums, and documentation.}
  \label{fig:pipeline}
\end{figure*}

All three datasets use the same target format: $256\times256$ object-centred crops with bounding-box annotations, split metadata, and release documentation. The source of the boxes differs across datasets. TrafficSigns uses YOLOv5x candidate boxes followed by manual review, Cityscapes--Pedestrian uses CityPersons and gtCoarse annotation sources, and COCO PottedPlant uses COCO instance annotations. Figure~\ref{fig:pipeline} shows how these different inputs are converted into a common release format.

The workflow has five stages: (1) detection or annotation lookup, (2) box extraction or conversion with class harmonization, (3) object-centred crop generation, (4) resizing to $256\times256$, and (5) filtering and split generation. The final bounding boxes can also be converted into simple binary masks when a model requires box-guided input.

The three datasets were deliberately selected to probe complementary constrained-data challenges rather than to equalize record counts. Cityscapes--Pedestrian stresses dense privacy-blurred scenes with occlusion. TrafficSigns provides cleaner high-contrast sign crops. COCO PottedPlant preserves context-rich indoor/outdoor variation and multi-instance crops. Across all subsets, $256\times256$ crops and bounding-box annotations provide a common interface from which masks can be rasterized when required.

For \textbf{Cityscapes--Pedestrian}, we begin from high-resolution Cityscapes frames \cite{cityscapes} and combine CityPersons bounding-box annotations \cite{citypersons} with gtCoarse annotations converted to bounding boxes where needed. Pedestrian-related labels such as \emph{pedestrian}, \emph{sitting person}, \emph{person group}, \emph{person (other)}, \emph{rider}, \emph{person}, \emph{persongroup}, and \emph{ridergroup} are treated as person-related categories and mapped into a unified pedestrian class for the released subset. Likely duplicate boxes across annotation sources are removed using overlap-based rules, and candidate $256\times256$ crops are generated around person-related regions. A subsequent filtering stage controls subset difficulty using object-count and overlap statistics before the selected samples are converted into YOLO-format labels and public manifest records.

For \textbf{TrafficSigns}, we start from author-owned $1024\times768$ street-view imagery. Candidate bounding boxes for \emph{Traffic Sign} and \emph{Damaged Traffic Sign} objects were generated using YOLOv5x \cite{yolov5} and manually reviewed before release. The released YOLO labels should therefore be interpreted as a reviewed annotation layer rather than raw detector output. These categories are merged into a single TrafficSigns class, and each sign-centered sample is standardized to $256\times256$. The current direct-release package contains 3,009 images organized in a YOLO-compatible train/validation/test structure with 2,256 training images, 451 validation images, and 302 test images.

This simplicity is deliberate rather than incidental. Samples generally contain one or a small number of visually clear signs, the bounding boxes are usually well defined, and the signs are typically high-contrast and free of heavy blur or clutter. Compared with Cityscapes--Pedestrian, the TrafficSigns subset therefore offers a cleaner, lower-density regime for evaluating edge fidelity, shape preservation, and fine-detail synthesis without strong confounding from heavy occlusion or background complexity.

For \textbf{COCO PottedPlant}, we start from MS-COCO 2017 annotations \cite{coco} and select the \emph{potted plant} category by name. The subset is constructed per annotated instance rather than per image: instances that are too small to support a stable $256\times256$ crop are skipped, and for each remaining object a crop window is sampled so that the selected bounding box lies fully inside the crop with a random margin. When multiple plants intersect the crop, their boxes are retained in the cropped labels so that visible target objects are not left unannotated. Because the source images come from mixed native resolutions and both indoor and outdoor scenes, this subset exposes wider background and object-appearance variability than the TrafficSigns data.

\paragraph{Quality control and processing records.}
Each subset records how the released samples were produced. TrafficSigns stores image and label paths, split membership, class mapping, and checksums. Cityscapes--Pedestrian and COCO PottedPlant store public manifests, box-level annotation tables, metadata, and reconstruction scripts rather than upstream image files. The COCO pipeline also keeps intersecting potted-plant boxes in cropped labels so that visible target objects are not left unlabelled. These records make the release easier to inspect and help users understand how preprocessing choices affect task difficulty.

\section{Dataset Collection and Regime Characterization}
\begin{table}[t]
  \caption{Summary of the current object-centric dataset release. TrafficSigns is directly released with images and YOLO labels. Cityscapes--Pedestrian and COCO PottedPlant are released through public manifests, public per-box annotation tables, metadata, and reconstruction assets because their source imagery has upstream redistribution constraints.}
  \label{tab:datasets}
  \footnotesize
  \setlength{\tabcolsep}{2pt}
  \renewcommand{\arraystretch}{1.00}
  \begin{tabularx}{\columnwidth}{@{}l c c >{\raggedright\arraybackslash}X@{}}
    \toprule
    Dataset & Records & Release mode & Main characteristics \\
    \midrule
    Cityscapes--Pedestrian & 2,156 & non-image & Pedestrian-related classes merged into pedestrian; $256\times256$ crops; privacy blur, overlap, group scenes, dense occlusion \\
    TrafficSigns & 3,009 & direct & Single TrafficSigns class; YOLO train/val/test release; $256\times256$ crops; lower-density signs, high contrast, limited occlusion \\
    COCO PottedPlant & 7,679 & scripts/metadata & MS-COCO 2017 potted plant selected by category name; per-instance $256\times256$ crops; intersecting plant boxes retained; indoor/outdoor context diversity \\
    \bottomrule
  \end{tabularx}
\end{table} 

The reported records have different meanings depending on release mode. For TrafficSigns, records correspond to directly released images and YOLO labels. For Cityscapes--Pedestrian and COCO PottedPlant, records correspond to public manifest entries that describe locally reconstructable samples without redistributing upstream-constrained image files. This distinction is intentional: it allows the resource to remain useful and verifiable while respecting the different redistribution conditions of the source datasets.

\paragraph{Cityscapes--Pedestrian.}
This subset is the most challenging regime in the collection. It contains dense pedestrian scenes in which people frequently overlap or appear in groups. Faces are blurred for privacy, which removes high-frequency features that would otherwise help photorealistic generation. As a result, the dataset combines occlusion, density, blur, and structural ambiguity, making it a challenging test case for low-data object-centric generation.

\paragraph{TrafficSigns.}
This subset provides a cleaner and more controlled regime than the pedestrian and potted-plant subsets. Samples contain visually clear traffic signs with sharp boundaries and high contrast, usually with less density and less occlusion than the pedestrian subset. Although some crops contain multiple signs, the visual structure is comparatively simple and the target objects are typically well defined. This makes the dataset useful for evaluating edge preservation, shape fidelity, and object detail without strong interference from cluttered backgrounds. In the public release, TrafficSigns is the only subset distributed directly with image files and YOLO-format labels.

\paragraph{COCO PottedPlant.}
This subset introduces a different kind of difficulty: semantic and contextual diversity. Crops come from indoor and outdoor scenes with varied source resolutions, object appearances, and backgrounds. Because crops are generated per annotated instance with random offsets, the dataset preserves contextual variation and object-placement diversity beyond the more uniform TrafficSigns setting. The reconstruction pipeline also retains intersecting potted-plant boxes in cropped labels, reducing the risk of visible but unlabeled target objects in multi-plant scenes.

\paragraph{Complementary failure modes.}
The three subsets are designed to expose different weaknesses in generative models. Cityscapes--Pedestrian tests whether a method can synthesize plausible object structure when the target is crowded, partially occluded, or degraded by privacy blur. TrafficSigns tests whether a method preserves crisp boundaries, simple geometry, and fine visual details in a cleaner setting. COCO PottedPlant tests whether a method can handle contextual diversity, variable object placement, and multi-instance ambiguity. A model that performs well on only one subset may therefore be exploiting a narrow visual regime, whereas consistent performance across all three suggests broader object-centric robustness.

\begin{table}[t]
  \caption{Manifest- and annotation-derived regime statistics for the current release. Records are sample-level entries, while boxes are target bounding boxes from the released or reconstructable annotations. Multi-box records contain more than one target box.}
  \label{tab:regime_stats}
  \footnotesize
  \setlength{\tabcolsep}{2pt}
  \renewcommand{\arraystretch}{1.0}
  \begin{tabularx}{\columnwidth}{@{}l r r r r@{}}
    \toprule
    Dataset & Records & Boxes & Boxes/rec. & Multi-box \\
    \midrule
    Cityscapes--Pedestrian & 2,156 & 17,479 & 8.11 & 100.0\% \\
    TrafficSigns & 3,009 & 4,607 & 1.53 & 51.3\% \\
    COCO PottedPlant & 7,679 & 20,230 & 2.63 & 59.7\% \\
    \bottomrule
  \end{tabularx}
\end{table}
These release-derived statistics provide a lightweight validation of the intended regime separation. Cityscapes--Pedestrian emphasizes dense pedestrian structure, TrafficSigns provides a cleaner lower-density sign regime, and COCO PottedPlant captures context-rich scenes with frequent multi-instance structure. Because the statistics are computed from public manifests and annotation tables, they can be recomputed from the release.
\begin{figure}[t]
  \centering
  \includegraphics[width=\columnwidth]{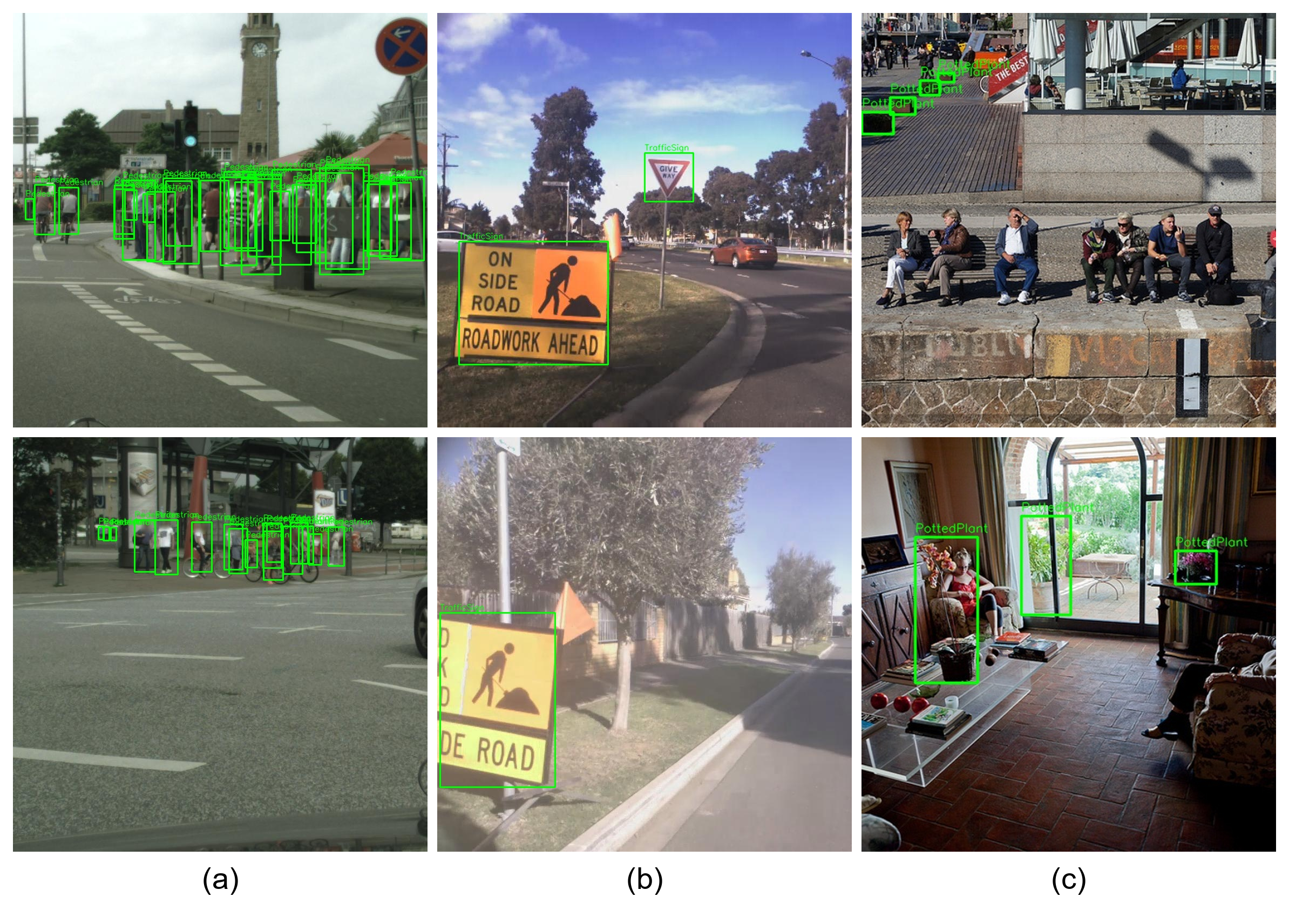}
\caption{Representative samples from the three regimes: dense Cityscapes--Pedestrian scenes, cleaner TrafficSigns crops, and context-diverse COCO PottedPlant crops.}
 \Description{A 2 by 3 grid of representative real 256 by 256 crops from the three released datasets: Cityscapes--Pedestrian examples with dense urban scenes, overlap, and privacy blur; TrafficSigns examples with clean isolated signs; and COCO PottedPlant examples with diverse indoor and outdoor backgrounds.}
  \label{fig:examples}
\end{figure}

\section{Research Applications, Release, and Conclusion}

The collection can be used for object-centric image generation, few-shot generation, synthetic-data augmentation, generative evaluation, and object-level representation learning. It is not tied to one model. The three subsets provide different visual regimes, and the release package provides shared metadata, bounding-box annotations, manifests, annotation tables, split documentation, and reconstruction scripts.

\paragraph{Resource validation and benchmarking use.}
A recent YOLOv11 evaluation used the TrafficSigns, Cityscapes--Pedestrian, and COCO PottedPlant regimes with real-only held-out test splits and fixed training settings. In the from-scratch setting, best synthetic augmentation improved Cityscapes--Pedestrian by +7.6\% and COCO PottedPlant by +30.6\%, while TrafficSigns showed a smaller +1.5\% gain consistent with its cleaner regime \cite{yolometrics}. Together with the release-derived statistics in Table~\ref{tab:regime_stats}, this prior use indicates that the collection can support studies of object structure, diversity, and detector performance across different visual regimes. The public manifests let users compare equal-size subsets with a fixed random seed, and the checksum tool helps users verify downloaded release files.

\paragraph{What is released.}
The release is hosted at \url{https://github.com/Latzi/object-centric-low-data-datasets} and archived on Zenodo at \url{https://doi.org/10.5281/zenodo.20573001}. TrafficSigns is distributed directly with images, YOLO labels, splits, metadata, checksums, a loader, and validation examples. Cityscapes--Pedestrian and COCO PottedPlant provide scripts, manifests, box-level annotation CSVs, metadata, split/mask documentation, pipeline configuration, checksums, and license notices, but not upstream-derived image files. For these two subsets, each manifest row describes one sample, and each annotation CSV row describes one bounding box. The reconstruction scripts and pipeline configuration explain how these records are used locally. Users can inspect the manifests, annotation tables, sampled subsets, checksum files, processing notes, ethics documentation, license notices, and citation metadata without reconstructing restricted images. The repository also includes tools for matched-size manifest sampling and checksum verification. These files let users inspect split membership, count boxes, verify file integrity, and build equal-size comparison sets before reconstructing any restricted images. The release can therefore be used either as a direct TrafficSigns dataset or as a reconstruction guide for the Cityscapes- and COCO-derived subsets.

\paragraph{Availability, licensing, and ethics.}
TrafficSigns is author-owned and released directly. Cityscapes--Pedestrian and COCO PottedPlant do not include original or cropped upstream images, thumbnails, previews, or derived mask images. Users must obtain Cityscapes and COCO separately and follow their terms. Responsible-use notes describe privacy, licensing, intended research use, and restrictions on surveillance, identity recognition, and attempts to bypass upstream terms.

\paragraph{Limitations.}
The collection is intentionally focused rather than comprehensive and does not replace large-scale generative datasets or general-purpose detection benchmarks. The mixed release model also means that TrafficSigns can be used directly, whereas Cityscapes--Pedestrian and COCO PottedPlant require local reconstruction from upstream data. This reduces convenience but preserves reproducibility while respecting redistribution constraints.

\paragraph{Conclusion.}
This paper presents a curated object-centric dataset collection for low- and constrained-data image generation and augmentation. The main contribution is a reusable release with a common crop-and-box format, three complementary visual regimes, release-derived statistics, direct data where allowed, public manifests, annotation tables, metadata, scripts, checksums, and reconstruction documentation.

\section*{Funding and Competing Interests}
\noindent
The authors report no competing interests. This work received no external funding.
% ----------------------------------------------------------------
% Required by the CIKM resource-paper call.
% Place this section immediately before the references.
% ----------------------------------------------------------------

\section*{GenAI Usage Disclosure}
\noindent
The authors used ChatGPT for identifying and correcting grammatical errors, typos, wording issues, formatting problems, and repository-documentation wording. In accordance with academic integrity requirements, all research questions, dataset construction decisions, methodology, release design, factual claims, analyses, and final manuscript content were developed, conducted, verified, and validated by the authors.

\end{document}